\documentclass[twoside,11pt]{article}

\usepackage{enumitem}
\usepackage{jmlr2e}
\usepackage{booktabs}
\usepackage{pifont} 
\usepackage{amsmath}
\usepackage{rotating}
\usepackage{cleveref}
\usepackage{float} 
\usepackage{tikz}
\usepackage{pgfplots}
\usepgfplotslibrary{groupplots}
\pgfplotsset{compat=newest}
\usetikzlibrary{external}

\pgfplotsset{%
  compat=newest,
  table/col sep=comma,
  mark end/.style={%
    scatter,
    scatter src=x,
    scatter/@pre marker code/.code={%
      \pgfmathtruncatemacro\usemark{%
        (\coordindex==(\numcoords-1))
      }
      \ifnum\usemark=0
      \pgfplotsset{mark=none}
      \fi
    },
    scatter/@post marker code/.code={}
  },
}

\usepackage{rotating}

\newcommand{\GPflow}{\mbox{GPf\hspace{1pt}low}}

\newcommand{\cmark}{\ding{51}}
\newcommand{\xmark}{\ding{55}}

\newlength\figureheight
\newlength\figurewidth

\newif\ifarxiv
\arxivtrue

\ifarxiv
  \jmlrheading{}{}{}{}{}{Matthews et al}
  \makeatletter
  \def\@starteditor{\noindent \small {}}
  \makeatother
\else
  \jmlrheading{}{2016}{}{}{}{Matthews et al}
\fi

\ShortHeadings{\GPflow: A Gaussian process library using TensorFlow}{Matthews et al}
\firstpageno{1}

\begin{document}

\title{\GPflow: A Gaussian process library using TensorFlow}

\author{\name Alexander G. de G. Matthews \email am554@cam.ac.uk\\
       \addr Department of Engineering\\
       University of Cambridge\\
       Cambridge, UK
       \AND
       \name Mark van der Wilk \email mv310@cam.ac.uk\\
       \addr Department of Engineering\\
       University of Cambridge\\
       Cambridge, UK
       \AND
       \name Tom Nickson \email tron@robots.ox.ac.uk\\
       \addr Department of Engineering Science\\
       University of Oxford\\
       Oxford, UK
       \AND
       \name Keisuke Fujii \email fujii@me.kyoto-u.ac.jp\\
       \addr Department of Mechanical Engineering and Science\\ 
       Graduate School of Engineering, Kyoto University, Japan
       \AND
       \name Alexis Boukouvalas \email alexis.boukouvalas@manchester.ac.uk\\
       \addr Division of Informatics\\ 
       Manchester University\\
       Oxford Road, Manchester, UK
       \AND
       \name Pablo Le\'on-Villagr\'a \email pablo.leon@ed.ac.uk\\
       \addr School of Informatics\\
       University of Edinburgh\\
       10 Crichton Street, Edinburgh, UK
       \AND
       \name Zoubin Ghahramani \email zoubin@eng.cam.ac.uk\\
       \addr Department of Engineering\\
       University of Cambridge\\
       Cambridge, UK
       \AND
       \name James Hensman \email james.hensman@lancaster.ac.uk\\
       \addr CHICAS, Faculty of Health and Medicine\\
       Lancaster University\\
       Lancaster, UK
}

\ifarxiv
  \editor{} 
\else
  \editor{TBA}
\fi

\maketitle
\ifarxiv
  \thispagestyle{plain} 
\fi

\begin{abstract}
\GPflow\ is a Gaussian process library that uses TensorFlow for its core computations and Python for its front end \footnote{\GPflow\ and TensorFlow are available as open source software under the Apache 2.0 license.}. The distinguishing features of \GPflow\ are that it uses variational inference as the primary approximation method, provides concise code through the use of automatic differentiation, has been engineered with a particular emphasis on software testing and is able to exploit GPU hardware.
\end{abstract}

\section{Existing Gaussian process libraries}

There are now many publicly available Gaussian process libraries ranging in scale from personal projects to major community tools. We will therefore only consider a relevant subset of the existing libraries. The influential GPML toolbox \citep{Rasmussen2010} uses \emph{MATLAB}. It has been widely forked. A key reference for our particular contribution is the GPy library \citep{GPy2014}, which is written primarily using Python and Numeric Python (NumPy). GPy has an intuitive object-oriented interface. Another relevant Gaussian process library is GPstuff \citep{Vanhatalo2013} which is also a \emph{MATLAB} library.

\section{Objectives for a new library}\label{section:flowObjectives}

\GPflow\ is motivated by a set of goals. The software is designed to be fast, particularly at scale. Where approximation inference is necessary we want it to be accurate. We aim to support a variety of kernel and likelihood functions. Another goal is that the implementations are verifiably correct. The software should be made easy to use by an intuitive user interface. Finally it should be easy to extend the software. We argue that there is a way to better meet these simultaneous objectives than existing packages and that it is realised in \GPflow. 

\section{Key features for meeting the objectives}\label{section:keyFlowFeatures}

To best meet the key goals of our library, we were led to a project that had all of the following distinguishing features:
\begin{enumerate}[label={(\roman*)}]
\item The use of variational inference as the primary approximation method to meet the twin challenges of non-conjugacy and scale. \label{feature:variational}
\item Relatively concise code which uses automatic differentiation to remove the burden of gradient implementations.\label{feature:autodiff}
\item The ability to leverage GPU hardware for fast computation.\label{feature:GPU}
\item A clean object-oriented Python front end. \label{feature:frontEnd}
\item A dedication to testing and open source software principles. \label{feature:goodPractice}
\end{enumerate}

Table \ref{table:featureComparison} gives a summary of which Gaussian process libraries possess the distinguishing features we have highlighted. The GPflow interface and Python architecture are heavily influenced by GPy. An important difference between \GPflow\ and GPy is that \GPflow\ uses TensorFlow for its core computations rather than numeric Python. This difference significantly affects the general requirements of the architecture. The GPU functionality in GPy is currently limited to CUDA code for the GPLVM \citep{Dai2014}. By contrast the \GPflow\ implementation is targeted at a broad variety of GPU capability. 

Having established a desirable set of key design features, the question arises as how best to engineer a Gaussian process library to achieve them. A central insight here is that many of the features we highlight are well supported in neural network libraries. Neural network software has made working with neural networks easier by using automatic differentiation to reduce the coding overhead for a user. Of the available libraries we use TensorFlow \citep{Abadi2015}, as discussed in the next section.
\begin{center}
\begin{table}[t]
\vskip-0.1in 
\caption{\label{table:featureComparison}A summary of the features possessed by existing Gaussian process libraries at the time of writing. OO stands for object-oriented. In the GPU column GPLVM denotes Gaussian process latent variable model and SVI is Stochastic variational inference. N\textbackslash R denotes not reported.}
\vskip 0.1in
\small
\begin{tabular}{lccccc}
\toprule
{\bf Library}  & {\bf Sparse variational} &{\bf Automatic} & {\bf GPU} &{\bf OO Python} &{\bf Test} \\
& {\bf inference} & {\bf differentiation} & {\bf demonstrated} & {\bf front end} & {\bf coverage} \\
\midrule
GPML & \cmark & \xmark & \xmark & \xmark & N\textbackslash R \\
GPstuff & Partial & \xmark & \xmark & \xmark & N\textbackslash R \\
GPy & \cmark & \xmark & GPLVM & \cmark &  49\% \\
\GPflow & \cmark & \cmark & SVI & \cmark & 99\%  \\
\bottomrule
\end{tabular}
\end{table}
\end{center}

\section{Contributing GP requirements to TensorFlow}

In TensorFlow \citep{Abadi2015} a computation is described as a directed graph where the nodes represent \emph{Operations} (\emph{Ops} for short) and the edges represent \emph{Tensors}. As a directed edge, a Tensor represents the flow of some data between computations. Ops are recognisable mathematical functions such as addition, multiplication etc. \emph{Kernels} (in an unfortunate collision in terminology with the Gaussian process literature) are implementations of a given Op on a specific device such as a CPU or GPU.  Like almost all modern neural network software, TensorFlow comes with the ability to automatically compute the gradient of an objective function with respect to some parameters. The TensorFlow open source implementation comes with a wealth of GPU kernels for the majority of Ops.

Although we gained significantly from using TensorFlow within GPflow, there were some capabilities that were not yet present in the software which were required for our purposes. We therefore added this functionality to TensorFlow.  Gaussian process software needs the ability to solve systems of linear equations using common linear algebra algorithms. The differentiation of computational graphs that used the Cholesky decomposition required a new Op, which we contributed with the help of Rasmus Munk Larsen, who reviewed the code. The main part of the code was a C++ implementation of the blocked Cholesky algorithm proposed by Murray \citeyearpar{Murray2016}. We also contributed code that enabled GPU solving of matrix triangular systems, which in some cases is the bottleneck for approximate inference.

\section{Details of \GPflow}\label{section:flowFunctionality}

\GPflow\ supports exact inference where possible, as well as a variety of approximation methods. One source of intractability is non-Gaussian likelihoods, so it is helpful to categorize the available likelihood functionality on this basis. Another major source of intractability is the adverse scaling of GP methods with the number of data points. To this end we support `variationally sparse' methods which ensure that the approximation is scalable and close in a Kullback-Leibler sense to the posterior \citep{Matthews2016}. Whether or not a given inference method uses variational sparsity is another useful way to categorize it. The inference options, which are implemented as classes in \GPflow, are summarized in Table \ref{table:flowModelClasses}. Note that all the MCMC based inference methods support a Bayesian prior on the hyperparameters, whereas all other methods assume a point estimate.  Our main MCMC method is Hamiltonian Monte Carlo (HMC) \citep{Neal2010}. 

\begin{table}[t]
\begin{center}
\vskip-0.1in 
\caption{\label{table:flowModelClasses} A table showing the inference classes in \GPflow. Relevant references are VGP \citep{Opper2009}, SGPR \citep{Titsias2009}, SVGP \citep{Hensman2013,Hensman2015} and SGPMC \citep{Hensman2015b}.}
\vskip 0.1in
\small
\begin{tabular}{lccc}
\toprule
{} & {\bf Gaussian }  & {\bf Non-Gaussian} &{\bf Non-Gaussian} \\
{} & {\bf likelihood} & {\bf likelihood} & {\bf likelihood} \\
{} & {} & {\bf (variational)} & {\bf (MCMC)} \\
\midrule
Full covariance & GPR & VGP & GPMC \\
Variational sparsity & SGPR & SVGP & SGPMC \\
\bottomrule
\end{tabular}
\end{center}
\vskip -0.2in
\end{table}

We now discuss some architectural considerations. The whole Python component of \GPflow\ is intrinsically objected-oriented. The code for the various inference methods in Table \ref{table:flowModelClasses} is structured in a class hierarchy, where common code is pulled out into a shared base class. The object-oriented paradigm, whilst very natural for the Python layer of the code, has a different emphasis to that of a computational graph which is arguably closer to a functional concept. The largely functional computational graph, and a object-oriented interface need to live cleanly together in GPflow. This is achieved through the Param class that allows parameters to be both properties of an object that can be manipulated as such and Variables in a TensorFlow graph that can be optimized.

A number of steps have been taken to ensure project quality and usability. All GPflow source code is openly available on GitHub at \url{http://github.com/GPflow/GPflow}.  The web page uses continuous integration to run an automated test suite. The test code coverage for GPflow is higher than similar packages where the code coverage statistics are published, achieving a level of $99\%$ (Table \ref{table:featureComparison}). A user manual can be found at \url{http://gpflow.readthedocs.io}. 

\section{Timed experiments}\label{section:gpflowExperiments}

As a scenario, we studied training a multiclass GP classifier on MNIST using stochastic variational inference \citep{Hensman2015,Hensman2015b} \footnote{Code for these timing experiments can be found at \url{http://github.com/alexggmatthews/GPflow_profiling}}. We compared against GPy. None of the other libraries discussed support this algorithm. Functionally, the algorithms are nearly identical in \GPflow\ and GPy.  We did a series of trials measuring the time each package took to perform $50$ iterations of the algorithm. The trials included a set of CPU experiments, where we varied the number of threads available to the two packages. For \GPflow, we also measured the effect of adding a GPU on top of the maximal number of CPU threads considered. GPy does not presently have a GPU implementation of this algorithm. We used a Linux workstation Intel Core I7-4930K CPU clocked at 3.40GHz an NVIDIA GM200 Geforce GTX Titan X GPU. 

\begin{figure*}
\ifarxiv
\centering\includegraphics{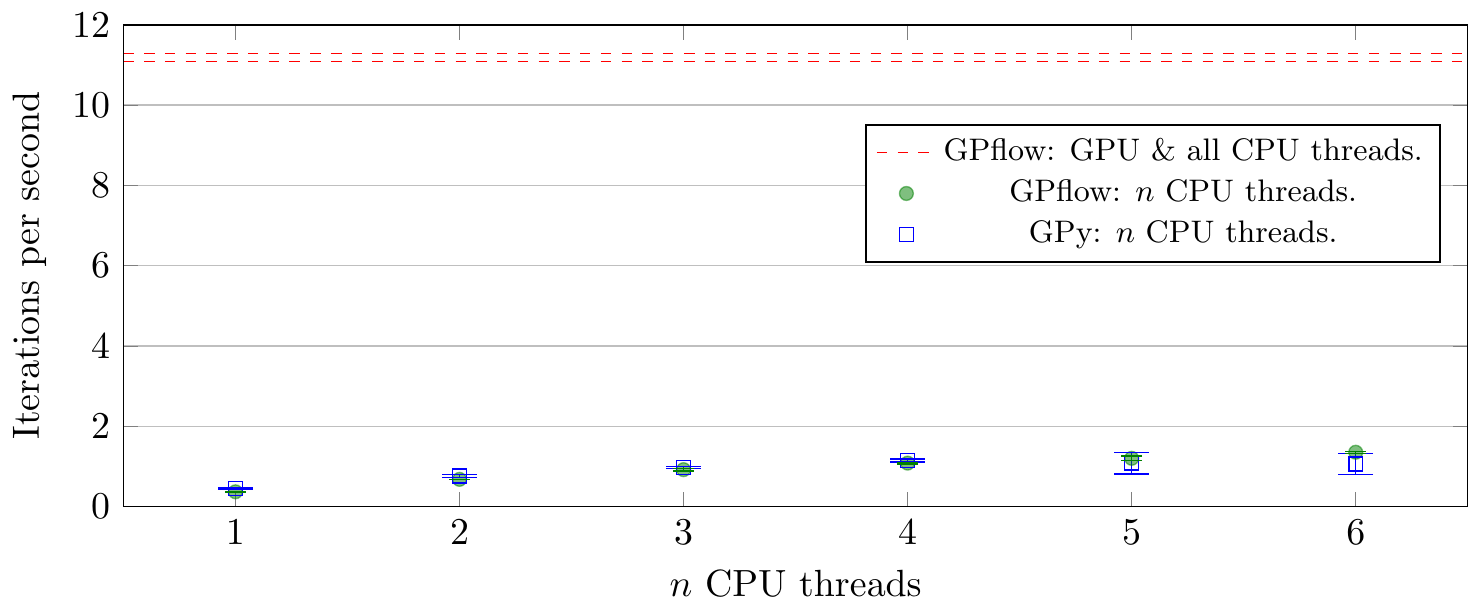}
\else
\setlength\figureheight{0.3\textheight}
\setlength\figurewidth{\textwidth}
\centering\input{GPflow_timing.tikz}
\fi
\caption{\label{fig:GPflowTiming}
A comparison of iterations of stochastic variational inference per second on the MNIST dataset for GPflow and GPy. Error bars shown represent one standard deviation computed from five repeats of the experiment.
}
\vskip -0.15in
\end{figure*}

The results of the timing experiments are shown in Figure \ref{fig:GPflowTiming}. For the CPU experiments, the speeds for \GPflow\ and GPy are similar. It can be seen that the increase in speed from adding a GPU is considerable. These gains could make a significant difference to the work flow of a researcher on this topic. Based on the measurements we have made, training using GPflow with $6$ CPU threads would take approximately $41$ hours or just under $2$ days. Adding a GPU would currently reduce the training time to about $5$ hours.

\section{Acknowledgements}

We acknowledge contributions from Valentine Svensson, Dan Marthaler, David J. Harris, Rasmus Munk Larsen and Eugene Brevdo. Alexander G. de G. Matthews was supported by EPSRC grants EP/I036575/1 and EP/N014162/1. James Hensman was supported by an MRC fellowship.

\tikzsetnextfilename{GPflow_timing}

\bibliography{bibliography.bib} 

\end{document}